\documentclass[lettersize,journal]{IEEEtran}
\usepackage{amsmath,amsfonts}
\usepackage{algorithmic}
\usepackage{algorithm}
\usepackage{array}
\usepackage[caption=false,font=normalsize,labelfont=sf,textfont=sf]{subfig}
\usepackage{textcomp}
\usepackage{stfloats}
\usepackage{url}
\usepackage{verbatim}
\usepackage{graphicx}
\usepackage{cite}
\usepackage{threeparttable}
\usepackage{tabularx}
\usepackage{makecell}
\usepackage{siunitx}
\usepackage{multirow} 
\usepackage{color}
 \usepackage{booktabs} 
 \usepackage{hyperref}
\begin{document}

\title{XGrasp: Gripper-Aware Grasp Detection with Multi-Gripper \\Data Generation}
\author{Yeonseo Lee$^{1}$,~Jungwook Mun$^{1}$,~Hyosup Shin$^{1}$,~Guebin Hwang$^{1}$,~%
        Junhee Nam$^{1}$,~Taeyeop Lee$^{1\dagger}$,~and~Sungho Jo$^{1\dagger}$%
\thanks{$^{1}$Korea Advanced Institute of Science and Technology (KAIST), Daejeon, South Korea.}%
\thanks{$^{\dagger}$Corresponding authors.}%
}



\maketitle

\begin{abstract}
Real-world robotic systems frequently require diverse end-effectors for different tasks, however most existing grasp detection methods are optimized for a single gripper type, demanding retraining or optimization for each novel gripper configuration. 
This gripper-specific retraining paradigm is neither scalable nor practical. 
We propose XGrasp, a real-time gripper-aware grasp detection framework that generalizes to novel gripper configurations without additional training or optimization.
To resolve data scarcity, we augment existing single-gripper datasets with multi-gripper 
annotations by incorporating the physical characteristics and closing trajectories of diverse grippers.
Each gripper is represented as a two-channel 2D image encoding its static shape (Gripper Mask) and dynamic closing trajectory (Gripper Path).
XGrasp employs a hierarchical two-stage architecture consisting of a Grasp Point Predictor (GPP) and an Angle-Width Predictor (AWP).
In the AWP, contrastive learning with a quality-aware anchor builds a gripper-agnostic embedding space, enabling generalization to novel grippers without additional training.
Experimental results demonstrate that XGrasp outperforms existing gripper-aware methods in both grasp success rate and inference speed across diverse gripper types.
Project page: \url{https://sites.google.com/view/xgrasp}
\end{abstract}

\section{Introduction}
\IEEEPARstart{R}{obotic} grasping is one of the most fundamental capabilities of autonomous manipulation systems and an essential component of a wide range of applications, from industrial automation to service robotics\cite{xie2023learning,marwan2021comprehensive}.
To meet this demand, a variety of end-effectors have been developed, including 2-finger parallel-jaw grippers\cite{mahler2019learning}, multi-finger hands\cite{li2022survey}, and task-specific grippers\cite{wang2022challenges}, each suited to different operational requirements.

Grasp detection has achieved remarkable progress, 
with 2D planar grasping being widely adopted in practice due to its 
simple and efficient\cite{kleeberger2020survey,caldera2018review,kumra2022gr,morrison2020learning}.
However, most existing methods are developed and optimized for a single gripper type\cite{kumra2017robotic,friedl2024evaluation}, resulting in a per-gripper model paradigm that is neither scalable nor generalizable: deploying a new gripper requires collecting dedicated training data and retraining from scratch.
Large-scale grasp datasets such as Cornell\cite{jiang2011efficient}, Jacquard\cite{depierre2018jacquard}, and Grasp-Anything\cite{vuong2024grasp} provide extensive grasp annotations, but are predominantly limited to the 2-finger parallel-jaw gripper.
While recent works such as MultiGripperGrasp\cite{casas2024multigrippergrasp} 
and GenDexGrasp\cite{li2023gendexgrasp} have introduced multi-gripper 
datasets, they are designed for grasp pose generation on isolated object 
inputs, which limits their scalability to practical robotic scenarios 
involving direct sensor inputs.

To address grasp detection across diverse gripper types, several gripper-aware methods have been proposed.
AdaGrasp\cite{xu2021adagrasp} encodes gripper geometry using Truncated Signed Distance Field(TSDF)\cite{newcombe2011kinectfusion}, 
HybGrasp\cite{mun2023hybgrasp} combines deep learning with reinforcement learning to capture gripper-specific characteristics, 
and HybridGen\cite{wang2024transferring} adopts a transfer learning framework with gripper-specific optimization for grasp prediction. 
Nevertheless, these approaches share a common limitation in terms of real-time applicability. 
AdaGrasp incurs high computational costs due to volumetric TSDF processing,
HybGrasp demands retraining for unseen grippers and is constrained by the overhead of reinforcement learning, and HybridGen requires substantial computation time due to its optimization process. 
Furthermore, methods for effectively acquiring multi-gripper training data and design generalizable gripper representations across diverse gripper types remains largely underexplored.

\begin{figure}
\centering
\includegraphics[width=0.95\columnwidth]{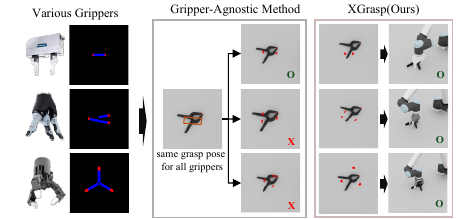}
\caption{Overview of XGrasp. While gripper-agnostic methods predict identical grasp poses regardless of gripper type, XGrasp generates gripper-aware grasp poses across diverse gripper configurations.}
\vspace{-15pt}
\label{fig:figure1}
\end{figure} 

To overcome these limitations, we propose XGrasp, a real-time 2D planar grasp detection framework that generalizes to diverse grippers without additional training or optimization.
First, to resolve data scarcity, we propose XG-Dataset, a multi-gripper grasp dataset constructed by augmenting existing single-gripper datasets. 
For each gripper, we extract a two-channel gripper representation encoding its static shape (Gripper Mask) and dynamic closing trajectory (Gripper Path), and apply a Graspability Decision Rule that evaluates collision, path intersection, and grasp stability to automatically generate valid grasp annotations across diverse gripper configurations.
Second, to achieve both real-time inference and high grasp success rate, XGrasp adopts a hierarchical two-stage architecture consisting of a Grasp Point Predictor (GPP) and an Angle-Width Predictor (AWP). 
In the AWP, contrastive learning with a quality-aware anchor builds an embedding space that captures structural distinctions between successful and failed grasps.
This allows the embedding space to remain valid for novel gripper configurations, enabling generalization to novel grippers without any additional fine-tuning.
Experimental results demonstrate that XGrasp outperforms existing gripper-aware methods in both inference speed and grasp success rate, showing that effective multi-gripper grasping is achievable without optimization or high-dimensional inputs.

The main contributions can be summarized as follows:
\begin{itemize}
    \item We propose a multi-gripper data augmentation methodology that automatically generates multi-gripper grasp annotations from existing single-gripper datasets.
    \item We design a two-stage hierarchical architecture that decouples grasp point prediction (GPP) and angle-width determination (AWP), simultaneously achieving real-time inference speed and high grasp success rate.
    \item We propose a quality-aware contrastive learning strategy for AWP to learn a gripper-agnostic embedding space anchored on optimal grasp samples, enabling zero-shot generalization to novel gripper configurations.
\end{itemize}

The remainder of this paper is organized as follows. Section \ref{section:Labeling} describes methods for augmenting existing grasp datasets with multi-gripper annotations. 
Section \ref{section:Method} explains the detailed structure and training methodology of the XGrasp framework, and Section \ref{section:Experiments} presents the experimental results of the proposed method.

\section{Related Works}
\subsection{Grasp Detection}
Planar grasp detection has progressed from early CNN-based regression methods~\cite{lenz2015deep, redmon2015real} to more advanced architectures.
Subsequent works adopted deeper residual networks~\cite{kumra2017robotic} to improve feature representation, while generative approaches~\cite{morrison2018closing, kumra2022gr} further advanced detection accuracy by directly predicting dense grasp maps in a single forward pass.
Throughout this progression, grasp representations have converged around the grasp rectangle formulation~\cite{jiang2011efficient}, which encodes grasp pose as a set of geometric parameters including position, angle, and width.
However, this representation remains implicitly optimized for a fixed gripper geometry and closing trajectory, posing a structural barrier to generalization across diverse gripper types.

More recently, Transformer-based attention mechanisms have been incorporated to capture long-range spatial dependencies in grasp detection~\cite{wang2022transformer, guo2024end, zhang2024effective}, further improving detection accuracy.
In parallel, large vision-language models and foundation model-based approaches have enabled language-conditioned grasping, allowing robots to detect grasp poses according to natural language instructions~\cite{noh2025graspsam, vuong2024language, 
nguyen2025graspmamba}.
Despite these advances, all of these methods remain constrained by grasp representations that presuppose a single gripper type, limiting their applicability to real-world scenarios requiring diverse end-effectors.
\subsection{Gripper-Aware Grasp Detection}
In real robotic systems, various grippers are deployed depending on task and object characteristics, yet existing grasp detection research has largely focused on single-gripper environments.
MultiGripperGrasp~\cite{casas2024multigrippergrasp} and GenDexGrasp~\cite{li2023gendexgrasp} have introduced multi-gripper datasets to address this data scarcity; 
however, both are designed for grasp pose generation on isolated object inputs, which limits their scalability to practical robotic applications.
To address gripper-aware grasp detection, several approaches have 
been proposed.
AdaGrasp~\cite{xu2021adagrasp} explicitly encodes gripper geometry as 3D TSDF volumes and feeds this high-dimensional representation throughout the network, enabling grasp prediction for diverse end-effectors.
However, this volumetric pipeline incurs substantial inference overhead, and deploying a new gripper still requires reconstructing its TSDF from a 3D CAD model.
HybGrasp~\cite{mun2023hybgrasp} adopts a two-stage architecture in which deep learning generates initial grasp candidates and reinforcement learning performs gripper-specific adaptation.
While this formulation allows flexible gripper-specific tuning, it fundamentally requires retraining the reinforcement learning stage for each new gripper, limiting scalability to unseen configurations.
HybridGen~\cite{wang2024transferring} combines learning-based grasp prediction with optimization-based transfer to propagate grasp knowledge across grippers, but the iterative optimization performed at inference time precludes real-time application.
Across these approaches, existing methods rely on at least one of three constraints: high-dimensional gripper representations, gripper-specific retraining, or gripper-specific optimization overhead. 
A real-time gripper-aware grasp detection framework that simultaneously overcomes all three limitations and generalizes to unseen grippers without additional training remains an open research challenge.


\section{Data Augmentation for Multiple Grippers}
\label{section:Labeling}
While existing grasp datasets have driven significant progress in grasp detection, most are designed for 2-finger parallel-jaw grippers\cite{CornellGraspingDataset,depierre2018jacquard,vuong2024grasp}.
This limits their ability to capture the varying kinematic constraints and morphological characteristics across diverse gripper types.
To overcome these constraints, we propose XG-Dataset, 
which augments existing single-gripper grasp datasets with multi-gripper annotations to enable training across diverse gripper configurations.
Rather than simply replicating data, it aims to achieve hardware universality by reinterpreting and extending existing single-gripper annotations to align with the physical characteristics of diverse end effectors. 
This process requires a methodology that abstracts hardware characteristics into a generalized representation, enabling the model to consistently handle grippers of different shapes and specifications. 
Therefore, this section describes the definition of the action space, the logical foundation for integrated control of diverse grippers, and the data generation process.

\subsection{Action Space and Width Discretization} 
To effectively handle grippers with different physical specifications within a single framework, it is first necessary to convert the complex movements of the gripper into a unified representation.  
To this end, this study defines a discrete action space $\mathcal{A}$ that abstracts the gripper's motion as a combination of angles and widths as follows:
$$\mathcal{A} = \{a_{i,j} = (\theta_i, j) \mid 0 \le i < N_a, 0 \le j < N_w\}$$
Here, $N_a$ denotes the number of grasp angle candidates, and $N_w$ represents the number of discrete widths defined in the action space. 
To completely reflect the morphological characteristics of the asymmetric gripper, the range of the grasping angle is set to the full $360^\circ$, and this is discretized at $5^\circ$ intervals, defining a total of $N_a = 72$ angle candidates ($i \in \{0, \dots, 71\}$). 
The specific angle corresponding to each index $i$ is calculated as $\theta_i = i \times 5^\circ$. 
The number of discrete widths is set to $N_w = 12$ ($j \in \{0, \dots, 11\}$).These values are chosen to provide sufficient resolution for practical grasp detection while maintaining computational efficiency.

Each gripper $g$ possesses its own physical opening range $[w_{min,g}, w_{max,g}]$. The width index $j \in {0, \dots, 11}$ predicted by the model is mapped to the actual physical width $W_g(j)$ of that gripper as follows.
$$W_g(j) = w_{min,g} + \frac{j}{N_w - 1} \times (w_{max,g} - w_{min,g})$$
This gripper-specific width normalization approach maximizes cross-gripper generalization performance by enabling the model to learn an abstracted common feature relative openness rather than relying directly on the absolute dimensions of each gripper.

\subsection{Gripper Mask: Static Geometry} 
Given a specific action $a_{i,j}$, the gripper mask $\mathcal{M}(a)$ is defined as the image projected onto a two-dimensional plane of the gripper fingertip's shape at the actual physical width $W_g(j)$ corresponding to width index $j$ and angle $\theta_i$. 
$$\mathcal{M}(a) = \{(u, v) \mid \mathbb{I}_{mask}(u, v; \theta_i, W_g(j)) = 1\}$$
Here, $\mathbb{I}_{mask}$ is a binary indicator function representing the gripper's fingertip region.

\subsection{Gripper Path: Dynamic Trajectory} 
The gripper path $\mathcal{P}(a)$ denotes the kinematic sweep trajectory traced by the fingers during the process of reaching the fully closed state ($w_{min,g}$) from the current open state ($W_g(j)$) determined by a specific action $a_{i,j}$, in order for the gripper to grasp an object.  It is defined as follows for the time parameter $t \in [0, 1]$.
$$\mathcal{P}(a) = \bigcup_{t \in [0, 1]} \mathcal{M}(\theta_i, w(t))$$
Here, $w(t)$ is a linear function that changes from the initial width $W_g(j)$ to the minimum width $w_{min,g}$. This path representation helps to precisely predict whether effective contact with the object is possible and potential interference by predefining the spatial occupancy region traversed by the gripper during actual operation.

\begin{figure}
\centering
\includegraphics[width=0.85\columnwidth]{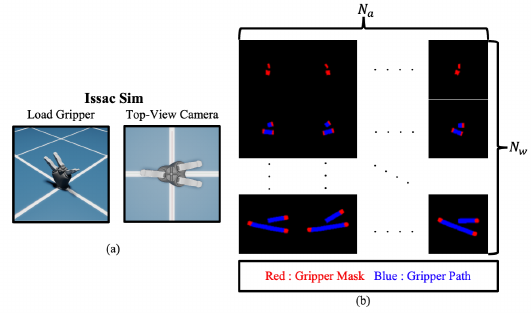}
\caption{Gripper Input Generation Process. (a) The target gripper is positioned facing a top-down camera in Isaac Sim to extract projected mask and path images for each angle-width combination. (b) Generated 2-channel gripper inputs across all $N_a \times N_w$ action combinations. Red: Gripper Mask, Blue: Gripper Path.} 
\vspace{-10pt}
\label{fig:figure2}
\end{figure} 
\subsection{Gripper Input Generation} \label{section:Gripper_Input}
This section describes the process of generating gripper features $G(a)$, the final input data for the learning model, by combining the previously defined gripper mask ($\mathcal{M}$) and gripper path ($\mathcal{P}$). 
The gripper feature $G(a)$ for a specific action $a$ is constructed by concatenating the mask, which represents static geometric information, and the path, which represents dynamic trajectory information, along the channel dimension. 
It is defined as follows:
$$G(a) = [\mathcal{M}(a), \mathcal{P}(a)] \in \{0, 1\}^{2 \times H \times W}$$

In this study, an automated pipeline utilizing the Isaac Sim simulation environment was constructed to extract Mask and Path information for various grippers efficiently. 
As shown in Fig. \ref{fig:figure2}-(a), the target gripper is positioned to face the top-down view camera directly, and projected images corresponding to each motion combination are acquired. 
Through this process, the following two-channel gripper features $\mathbf{G}(a)$ are generated for all angle-width combinations ($N_a \times N_w$):

\begin{itemize}
\item \textbf{Gripper Mask (Red channel)}: The static geometric shape of the gripper at a specific $\theta_i$ and  $W_g(j)$. 
\item \textbf{Gripper Path (Blue channel)}: The trajectory from the current state to the fully closed state.
\end{itemize}

Fig. \ref{fig:figure2}-(b) shows an example of the generated 2-channel input $G(a)$. This projection-based representation offers the advantage of being computationally more efficient than methods directly using high-dimensional data like 3D TSDF, while still sufficiently reflecting the gripper's unique shape and dynamic characteristics.

\subsection{Labeling Method for XG-Dataset}
This section describes how to generate annotations for the XG-Dataset using the gripper inputs created at an earlier stage.
We used the Jacquard Dataset\cite{depierre2018jacquard} as a base to build the XG-Dataset. 
The Jacquard Dataset provides 54,000 RGB-D scenes and 11 million grasp labels for parallel-jaw grippers. 
In this study, we used the grasp point  $(x, y)$ from the grasp labels $g = \{x, y, w, h, \theta\}$. 
\begin{figure}
\centering
\includegraphics[width=0.75\columnwidth]{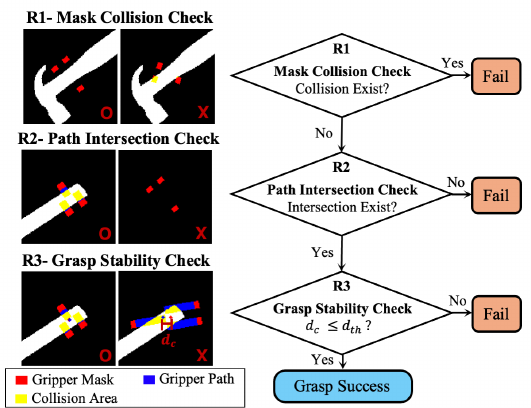}
\caption{Graspability Decision Rule consisting of three sequential checks: R1 Mask Collision Check, R2 Path Intersection Check, and R3 Grasp Stability Check.} 
\label{fig:figure3-1}
\vspace{-10pt}
\end{figure} 

\begin{figure}
\centering
\includegraphics[width=0.75\columnwidth]{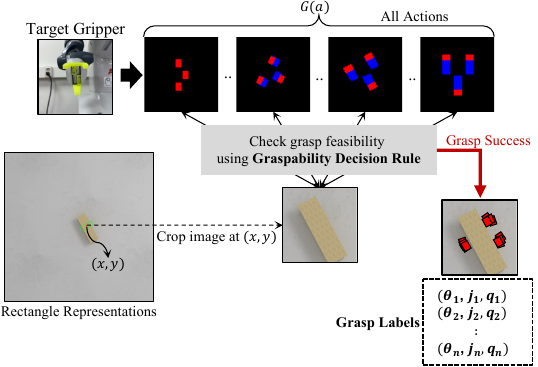}
\caption{The overall pipeline for generating target gripper grasp annotations.} 
\label{fig:figure3-2}
\vspace{-10pt}
\end{figure} 

To evaluate the grasping feasibility of different grippers, we extracted two-channel gripper features ${G}(a)$ of each target gripper using the pipeline described in section \ref{section:Gripper_Input}.
The overall pipeline for generating the target gripper's grasp annotations is shown in Fig. \ref{fig:figure3-2}. 
First, we crop the image around the grasp point $(x, y)$ from the Jacquard Dataset to define the analysis region.
Next, we check grasp feasibility using the Graspability Decision Rule (Fig. \ref{fig:figure3-1}).
The collision check (R1) evaluates the possibility of a collision between the gripper and the object by checking for overlap between the gripper mask and the object mask. 
If a collision is detected, the corresponding grasp is classified as a failure.
Then, the path intersection check (R2) evaluates whether the gripper's path intersects with the target object during the grasping motion. 
If no intersection is detected, the grasp is classified as a failure since the gripper cannot make contact with the object.
Finally, the grasp stability check (R3) compares the distance between the gripper's center point and the center point of the area where the gripper's path and the object intersect to determine whether the grasping position is stable.

For grasp candidates that pass all evaluation stages, we generate a final label consisting of a cropped scene image, grasp angle ($\theta_i$), grasp width index ($j$), and grasp quality ($q$).
In this study, we define the grasp quality based on the gripper width, prioritizing configurations that achieve a successful grasp with minimal finger opening. This is because a smaller grasp width indicates a more precise and stable interaction between the gripper and the object, minimizing potential slippage and unintended collisions with the environment. 
In our proposed XGrasp model, the structure simultaneously processes features $G(a)$ for all grasp action candidates at a specific grasp point. 
Reflecting this structural characteristic, we define quality $q$ based on the relative rank among candidates within a specific grasp point, rather than using global normalization.
When defining the $\mathcal{S}$ as the set of all valid grasp action candidates that pass all evaluation rules (R1–R3), the quality score $q(i, j)$ for each action is calculated as follows.
$$q(i, j) = 
\begin{cases} 
1, & \text{if } |\mathcal{S}| = 1 \\
1 - \frac{\text{rank}(j) - 1}{|\mathcal{S}| - 1}, & \text{if } |\mathcal{S}| > 1
\end{cases}$$
where the term $|\mathcal{S}|$ represents the total count of these feasible candidates, 
and $\text{rank}(j)$ indicates the ordinal position of width index $j$ when the valid width indices in $\mathcal{S}$ are sorted in ascending order.
Candidates sharing the same width index are assigned the same rank.
By mapping the smallest width among multiple feasible candidates to the highest quality score, the model is trained to identify the most efficient and refined grasp pose from a diverse set of grasp candidates. Consequently, this relative evaluation method enables the derivation of grasping strategies optimized for each situation, independent of the object's absolute size.

Using this method, we constructed the XG-Dataset for the five different types of grippers (2f-v1, 3f-v1, 3f-v2, 4f-v1, 4f-v2) presented in Fig. \ref{fig:figure9}. 
By generating grasp labels that reflect the unique physical characteristics and actuation mechanisms of each gripper, we were able to create a versatile training dataset applicable to various robotic grippers.

\section{Proposed Method}\label{section:Method}
\begin{figure*}
\centering
\includegraphics[width=\textwidth]{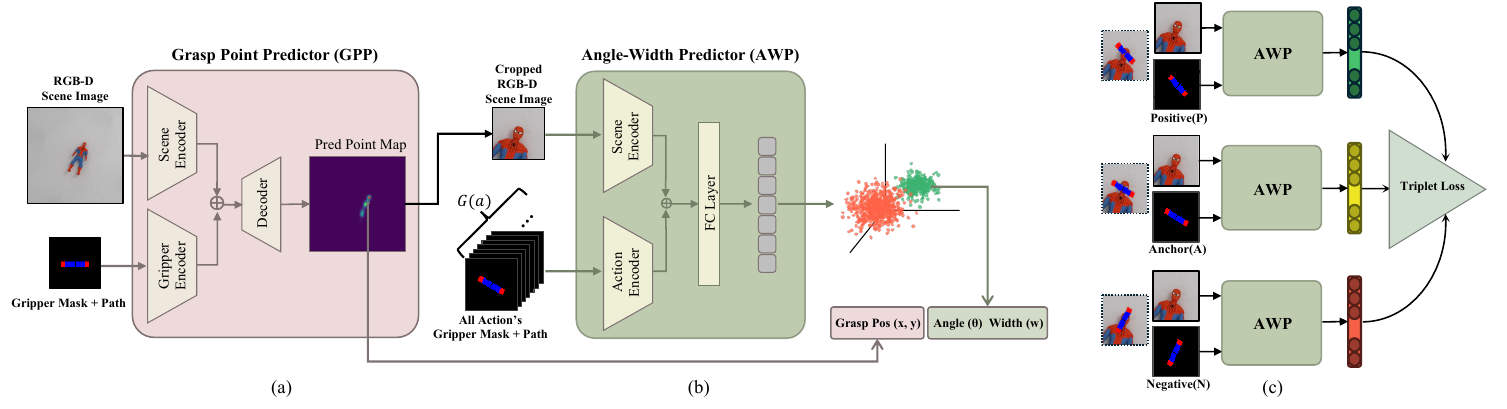}
\caption{Overview of the XGrasp framework: 
(a) Grasp Point Predictor (GPP) localizes the optimal grasp point from the full scene image and gripper input, 
(b) Angle-Width Predictor (AWP) determines the grasp angle and width from a cropped scene patch, and 
(c) AWP training with triplet loss and a quality-aware anchor to build an embedding space that generalizes across gripper types.}
\label{fig:figure4}
\vspace{-10pt}
\end{figure*} 
\subsection{Overview of XGrasp System}
We propose XGrasp, a two-stage approach for gripper-aware grasp detection.
The XGrasp system is composed of a two-stage architecture, as shown in Fig. \ref{fig:figure4}. 
In the first stage, the Grasp Point Predictor (GPP) utilizes the full scene image and the gripper input described in the previous section to predict a grasp point $(x, y)$.
In the second stage, the Angle-Width Predictor (AWP) receives a cropped scene image centered on the grasp point predicted by the GPP, along with the gripper inputs ($G(a)$) for all actions, and extracts an embedding for each action. 
The optimal grasp angle and width are derived from these embeddings.
In summary, the GPP is responsible for predicting a grasp point based on the global information of the scene, while the AWP utilizes local information around the grasp point to predict a precise grasp angle and width.

\subsection{Grasp Point Predictor (GPP)}
The Grasp Point Predictor(GPP) is a U-Net-based\cite{ronneberger2015u} model that takes a 4-channel RGB-D scene image and a 2-channel gripper input ($G(a)$) as inputs and outputs a grasp probability heatmap at the original resolution (224×224), as illustrated in Fig. \ref{fig:figure4}-(a). 
Scene and gripper features are extracted through separate encoder branches and fused channel-wise, allowing the decoder to localize graspable regions with awareness of both object geometry and gripper configuration.
The predicted heatmap is then used to extract the grasp point (x,y) as the location of maximum activation, and a scene patch is cropped centered on (x,y) and passed to the AWP for fine-grained angle and width determination.

\subsection{Angle-Width Predictor (AWP)}
The second stage, Angle-Width Predictor (AWP), takes the cropped scene and gripper inputs $G(a)$ for all $N_a \times N_w$ action candidates simultaneously.
For each action, the scene patch and the corresponding gripper input are concatenated along the channel dimension and encoded into an embedding vector in parallel, allowing the model to evaluate all possible grasp configurations in a single forward pass without iterative inference.
The AWP utilizes contrastive learning to learn the characteristics of successful and failed actions, enabling effective operation even with unseen grippers.
The detailed training process of the AWP model is presented in Fig. \ref{fig:figure4}-(c). 
\begin{figure}
\centering
\includegraphics[width=0.8\columnwidth]{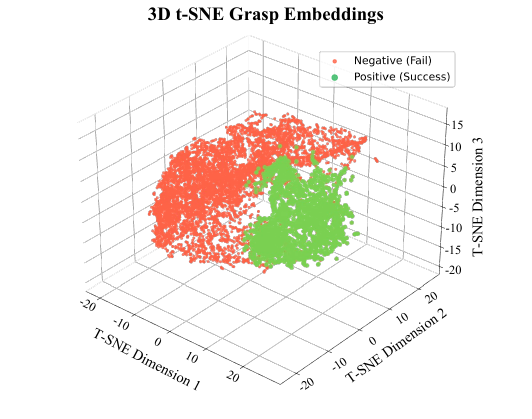}
\caption{3D t-SNE visualization of the 128-dimensional grasp embeddings for novel gripper configurations unseen during training.} 
\label{fig:figure6}
\vspace{-10pt}
\end{figure} 
Three scene-action pairs consisting of Anchor, Positive (successful grasp), and Negative (failed grasp) are input into a Siamese network\cite{koch2015siamese} with shared weights and encoded into 128-dimensional embedding vectors. 
Rather than simply distinguishing success from failure, the anchor is set to the action with the highest quality score ($q$) among successful candidates, guiding the model to densely cluster high-quality grasp regions in the embedding space. 
The triplet loss trains the network to minimize the distance between anchor and positive embeddings while maximizing the distance between anchor and negative embeddings in the learned embedding space:
\begin{equation}
\begin{split}
    L(A, P, N) = \max(&||f(A) - f(P)||^2 \\
                     & \quad - ||f(A) - f(N)||^2 + \alpha, 0)
\end{split}
\end{equation}
where $A$, $P$, and $N$ represent anchor, positive, and negative samples, respectively, $f(\cdot)$ denotes the embedding function, and $\alpha$ is the margin parameter.
Fig. \ref{fig:figure6} shows the 3D t-SNE\cite{maaten2008visualizing} visualization results of the 128-dimensional grasp embeddings obtained from novel grippers unseen during training. 
Each point corresponds to an individual grasp action $G(a)$, and it can be observed that failed (negative, red) and successful (positive, green) grasp embeddings form clearly distinguishable clusters. 
This demonstrates that the learned feature space has high discriminative performance.
This generalization is enabled by both the input structure and the physics-based nature of the training labels. 
By taking scene-action pairs as input, the AWP explicitly conditions its prediction on the gripper geometry and closing trajectory encoded in G(a), rather than implicitly assuming a fixed gripper type. 
Furthermore, rather than learning gripper-specific appearances, the AWP learns physical interaction features such as whether the gripper mask collides with the object and whether the gripper path intersects it, derived from the Graspability Decision Rule. 
Since these physical principles are invariant across gripper morphologies, the learned embedding space remains valid for novel gripper configurations without additional fine-tuning.

\section{Experiments} \label{section:Experiments}
In this section, we evaluate the performance of the proposed XGrasp through three sets of experiments:
(1) standard benchmark experiments on the Jacquard dataset to evaluate performance against existing methods,
(2) simulation experiments to verify generalization to novel grippers,
(3) real-world experiments to validate zero-shot generalization to physical robot systems.
Notably, all three experiments are conducted using a single XGrasp model without any additional training or optimization.
Additionally, we conducted ablation studies on the XG-Dataset composition and gripper input configuration.
Performance is measured by Success Rate (SR), defined as the number of successful grasps divided by the total number of attempts, reported both per gripper type and as overall average.

The following baseline models were utilized as comparison groups in the experiments:
\begin{itemize}
\item \textbf{GR-ConvNet\cite{kumra2020antipodal}}: 
Single-gripper method without gripper-type awareness 
\item \textbf{HybGrasp\cite{mun2023hybgrasp}}:
Multi-gripper method based on a hybrid deep learning and reinforcement learning architecture.
\item \textbf{HybridGen\cite{wang2024transferring}}: Multi-gripper method with additional grasp optimization for novel grippers.
\item \textbf{GR-ConvNet\cite{kumra2020antipodal}+AWP}: GR-ConvNet as grasp point predictor combined with the proposed AWP.
\item \textbf{FastSAM\cite{zhao2023fast}+AWP}: FastSAM as grasp point predictor combined with the proposed AWP.
\end{itemize}

\subsection{Jacquard Dataset Experiments} \label{section:Experiments_Jacquard}
\begin{table*}[t]
\centering
\caption{Performance comparison of grasping success rate and inference time on the Jacquard dataset}
\label{tab:table1}
\renewcommand{\arraystretch}{0.85}
\resizebox{\textwidth}{!}{%
\begin{tabular}{l *{14}{c} | *{2}{c}}
\toprule
\multirow{2}{*}{\textbf{Methods}} & \multicolumn{2}{c}{\textbf{WSG50}} & \multicolumn{2}{c}{\textbf{Franka}} & \multicolumn{2}{c}{\textbf{Robotiq-3F}} & \multicolumn{2}{c}{\textbf{Barrett}} & \multicolumn{2}{c}{\textbf{Kinova}} & \multicolumn{2}{c}{\textbf{Delto}} & \multicolumn{2}{c}{\textbf{DH3}} & \multicolumn{2}{c}{\textbf{Avg.}} \\
\cmidrule(lr){2-3} \cmidrule(lr){4-5} \cmidrule(lr){6-7} \cmidrule(lr){8-9} \cmidrule(lr){10-11} \cmidrule(lr){12-13} \cmidrule(lr){14-15} \cmidrule(l){16-17}
 & \textbf{SR} & \textbf{Time} & \textbf{SR} & \textbf{Time} & \textbf{SR} & \textbf{Time} & \textbf{SR} & \textbf{Time} & \textbf{SR} & \textbf{Time} & \textbf{SR} & \textbf{Time} & \textbf{SR} & \textbf{Time} & \textbf{SR $\uparrow$} & \textbf{Time $\downarrow$} \\
\midrule
GR-ConvNet\cite{kumra2020antipodal}   & 83.9 & 7   & 60.9 & 6   & 67.8 & 6   & 73.5 & 6   & 77.0 & 6   & 75.8 & 6   & 78.1 & 6   & 73.9 & 6.1  \\
HybGrasp\cite{mun2023hybgrasp}        & 83.9 & 272 & 74.7 & 259 & 86.2 & 257 & 81.6 & 264 & 82.7 & 257 & 86.2 & 257 & 78.1 & 269 & 81.9 & 262  \\
HybridGen\cite{wang2024transferring}  & 88.5 & 4750 & 74.7 & 4720 & 80.4 & 10570 & 79.3 & 10330 & 86.2 & 10740 & 88.5 & 8626 & 83.9 & 8600 & 83.1 & 8334 \\
GR-ConvNet+AWP                        & 86.2 & 25  & 82.7 & 26  & 88.5 & 28  & 80.4 & 26  & 83.9 & 27  & 85.0 & 26  & 80.4 & 26  & 83.9 & 26.3 \\
FastSAM\cite{zhao2023fast}+AWP        & 92.5 & 26  & 80.0 & 23  & 92.5 & 24  & 92.5 & 25  & 86.2 & 29  & 91.2 & 24  & 88.7 & 23  & 89.1 & 24.9 \\
\textbf{XGrasp (GPP+AWP)}               & \textbf{95.4} & 23 & \textbf{88.5} & 25 & \textbf{87.3} & 23 & \textbf{89.6} & 24 & \textbf{87.3} & 23 & \textbf{93.1} & 25 & \textbf{90.8} & 23 & \textbf{90.3} & 23.7 \\
\bottomrule
\end{tabular}%
}
\vspace{-10pt}
\end{table*}
\begin{figure}
\centering
\includegraphics[width=0.75\columnwidth]{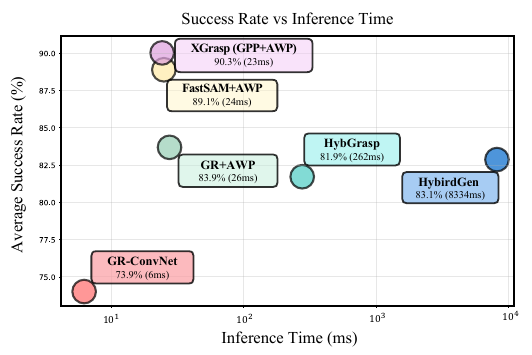}
\caption{Comparison with baseline methods on the Jacquard dataset.} 
\label{fig:figure7}
\vspace{-10pt} 
\end{figure} 
Following the standard evaluation protocol of the Jacquard dataset, all baseline methods were evaluated under identical experimental settings using seven novel gripper configurations (WSG50, Franka, Robotiq-3F, Barrett, Kinova, Delto, DH3), as shown in Fig. \ref{fig:figure_obj}-(b).
As shown in Table \ref{tab:table1}, the proposed method XGrasp achieves superior performance across all gripper types, with an average success rate of 90.3\%.
Although our method requires higher inference time than GR-ConvNet (6.1ms),
which does not consider gripper types, it significantly outperforms other gripper-aware methods in computational efficiency.
Specifically, the proposed method is over 10$\times$ faster than HybGrasp and 350$\times$ faster than HybridGen.
Fig. \ref{fig:figure7} presents a scatter plot of success rate versus inference time, demonstrating that the proposed method outperforms all gripper-aware baseline models in both metrics simultaneously.

\subsection{Simulation Experiments}
\begin{figure}
\centering
\includegraphics[width=0.75\columnwidth]{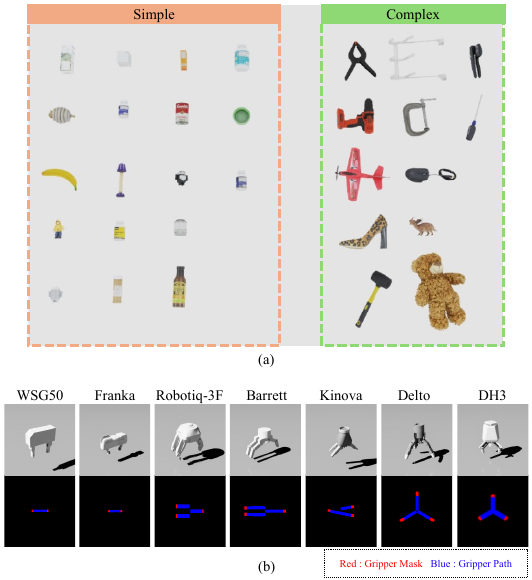}
\caption{Objects (a) and grippers (b) used in the simulation experiments} 
\label{fig:figure_obj}
\vspace{-10pt}
\end{figure}

The simulation experiments\cite{isaacsim2021} are designed to verify zero-shot generalization of XGrasp to novel gripper configurations unseen during training.
We selected 30 objects in total, comprising 18 with simple geometric shapes and 12 with complex shapes,  from the YCB Object\cite{calli2015ycb} and Google Scanned Objects\cite{downs2022google} datasets, as shown in Fig. \ref{fig:figure_obj}-(a).
A total of 5 grasp attempts were performed for each object, resulting in 90 trials for the simple objects and 60 trials for the complex objects. 
To validate generalization to novel grippers, seven gripper types unseen during training, shown in Fig. \ref{fig:figure_obj}-(b), were used in the experiment.

\begin{table*}[t]
\centering
\caption{Performance comparison of grasping success rates in simulation. S: Simple Object, C: Complex Object.}
\label{tab:table2}
\resizebox{\textwidth}{!}{%
\begin{tabular}{l *{14}{c} | *{3}{c}} 
\toprule
\multirow{2}{*}{\textbf{Method}} & \multicolumn{2}{c}{\textbf{WSG50}} & \multicolumn{2}{c}{\textbf{Franka}} & \multicolumn{2}{c}{\textbf{Robotiq-3F}} & \multicolumn{2}{c}{\textbf{Barrett}} & \multicolumn{2}{c}{\textbf{Kinova}} & \multicolumn{2}{c}{\textbf{Delto}} & \multicolumn{2}{c}{\textbf{DH3}} & \multicolumn{3}{c}{\textbf{Average $\uparrow$}} \\ 
\cmidrule(lr){2-3} \cmidrule(lr){4-5} \cmidrule(lr){6-7} \cmidrule(lr){8-9} \cmidrule(lr){10-11} \cmidrule(lr){12-13} \cmidrule(lr){14-15} \cmidrule(l){16-18}
 & S & C & S & C & S & C & S & C & S & C & S & C & S & C & \textbf{Simple} & \textbf{Complex} & \textbf{Total} \\ \midrule
GR-ConvNet\cite{kumra2020antipodal} & \textbf{87.8} & 75.0 & 65.6 & 60.0 & 85.6 & 65.0 & 70.0 & 65.0 & 64.4 & 65.0 & 82.2 & 63.3 & 63.3 & 53.3 & 74.1 & 63.8 & 69.0 \\
HybGrasp\cite{mun2023hybgrasp} & 84.4 & 71.7 & 63.3 & 61.7 & \textbf{88.9} & 78.3 & 75.6 & 68.3 & \textbf{83.3} & \textbf{81.7} & 83.3 & 73.3 & \textbf{82.2} & 70.0 & 80.1 & 72.1 & 76.1 \\
HybridGen\cite{wang2024transferring} & 81.1 & \textbf{85.0} & 64.4 & 65.0 & 85.6 & 71.7 & 77.8 & 66.7 & 75.6 & 75.0 & 77.8 & 70.0 & 75.6 & 70.8 & 76.8 & 72.0 & 74.4 \\
GR-ConvNet+AWP & 83.3 & 78.3 & 72.2 & 58.3 & 87.8 & 76.7 & 72.2 & 65.0 & 80.0 & 78.3 & 81.1 & 73.3 & 78.9 & 66.7 & 79.4 & 70.9 & 75.2 \\
FastSAM\cite{zhao2023fast}+AWP & 84.4 & 73.3 & 71.1 & 60.0 & 87.8 & 75.0 & 74.4 & 68.3 & 82.2 & 78.3 & 83.3 & 73.3 & 80.0 & 65.0 & 80.5 & 70.5 & 75.5 \\
\textbf{XGrasp (GPP+AWP)} & \textbf{87.8} & \textbf{85.0} & \textbf{73.3} & \textbf{68.3} & 87.8 & \textbf{80.0} & \textbf{82.2} & \textbf{71.7} & 82.2 & \textbf{81.7} & \textbf{88.9} & \textbf{80.0} & \textbf{82.2} & \textbf{71.7} & \textbf{83.5} & \textbf{76.9} & \textbf{80.2} \\ \bottomrule
\end{tabular}%
}
\vspace{-10pt}
\end{table*}
As shown in Table \ref{tab:table2}, the proposed method achieves the best overall average success rate of 80.2\% across all comparison models.
GR-ConvNet recorded the lowest performance at 69.0\%, as its gripper-agnostic design does not consider gripper-specific geometric constraints, suggesting that incorporating gripper characteristics is essential for robust performance across diverse gripper types.
Furthermore, while FastSAM+AWP achieved 80.5\% on simple objects, comparable to our method, its performance dropped to 75.5\% on complex objects.
This performance gap can be attributed to the fact that FastSAM, as a general segmentation model, simply sets the object centroid as the grasp point, which proves insufficient for objects with complex geometries.

\begin{figure}
\centering
\includegraphics[width=\columnwidth]{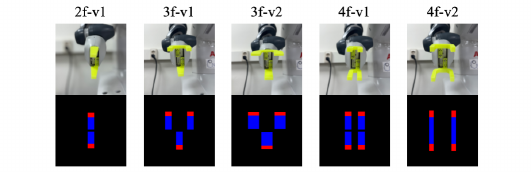}
\caption{Grippers used in real-world experiments.} 
\label{fig:figure9}
\vspace{-10pt}
\end{figure} 

\subsection{Real-World Experiments}
\begin{table*}[t]
\centering
\caption{Performance comparison of grasping success rates in a real-world environment. S: Simple Object, C: Complex Object}
\label{tab:table3}
\renewcommand{\arraystretch}{0.85}{%
\begin{tabular}{l *{10}{c} | *{3}{c}}
\toprule
\multirow{2}{*}{\textbf{Method}} & \multicolumn{2}{c}{\textbf{2f-v1}} & \multicolumn{2}{c}{\textbf{3f-v1}} & \multicolumn{2}{c}{\textbf{3f-v2}} & \multicolumn{2}{c}{\textbf{4f-v1}} & \multicolumn{2}{c}{\textbf{4f-v2}} & \multicolumn{3}{c}{\textbf{Average $\uparrow$}} \\
\cmidrule(lr){2-3} \cmidrule(lr){4-5} \cmidrule(lr){6-7} \cmidrule(lr){8-9} \cmidrule(lr){10-11} \cmidrule(l){12-14}
 & S & C & S & C & S & C & S & C & S & C & \textbf{Simple} & \textbf{Complex} & \textbf{Total} \\ \midrule

GR-ConvNet\cite{kumra2020antipodal} & 88.9 & 53.3 & 68.9 & 66.7 & 68.9 & 60.0 & 71.1 & 46.7 & 66.7 & 46.7 & 72.9 & 54.7 & 68.3 \\
HybGrasp\cite{mun2023hybgrasp} & 86.7 & 73.3 & 84.4 & 73.3 & 80.0 & \textbf{80.0} & 86.7 & \textbf{73.3} & \textbf{86.7} & 73.3 & 84.9 & 73.3 & 82.0 \\
HybridGen\cite{wang2024transferring} & 86.7 & 80.0 & 86.7 & 73.3 & 82.2 & 73.3 & 86.7 & 66.7 & 82.2 & 66.7 & 84.9 & 72.0 & 81.7 \\ 
\textbf{XGrasp (GPP+AWP)} & \textbf{91.1} & \textbf{86.7} & \textbf{88.9} & \textbf{93.3} & \textbf{91.1} & \textbf{80.0} & \textbf{91.1} & \textbf{73.3} & \textbf{86.7} & \textbf{80.0} & \textbf{89.8} & \textbf{82.7} & \textbf{88.0} \\ \bottomrule

\end{tabular}%
}
\vspace{-11pt}
\end{table*}

\begin{figure}
\centering
\includegraphics[width=0.8\columnwidth]{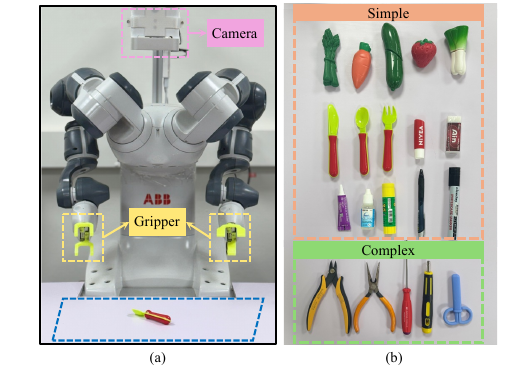}
\caption{The setup for the real-world grasping experiment. (a) the overall experimental setup and (b) the objects used for the grasping tasks.} 
\label{fig:figure10}
\vspace{-10pt}
\end{figure} 
The real-world experiments are designed to validate zero-shot cross-domain generalization of XGrasp to physical robot systems, conducted using an ABB IRB 14000 Yumi robot (7-DOF dual arm) and a Microsoft Azure Kinect RGB-D camera, as shown in Fig. \ref{fig:figure10}-(a).
We evaluated performance on 20 objects in total, comprising 15 simple household objects and 5 complex tool objects shown in Fig. \ref{fig:figure10}-(b), across five gripper types (2f-v1, 3f-v1, 3f-v2, 4f-v1, 4f-v2) shown in Fig. \ref{fig:figure9}, with three grasp attempts per object for a total of 60 trials.
As summarized in Table \ref{tab:table3}, XGrasp achieves the best performance across all gripper types with an average success rate of 88.0\%, demonstrating consistent improvements over all baseline models for both simple and complex objects under real-world conditions, including sensor noise and physical uncertainty.

\subsection{Ablation Study}
We conduct ablation studies to analyze the contribution of three key design choices: (1) multi-gripper data augmentation, (2) gripper input configuration (Mask and Path), and (3) AWP loss function design. All ablation studies follow the same experimental protocol as described in Section \ref{section:Experiments_Jacquard}.

\subsubsection{Importance of Data Augmentation for Multi-Gripper}


\begin{table*}[ht]
\centering
\footnotesize
\caption{Ablation study on training dataset composition. (1): Jacquard Dataset, (2)$\sim$(4): XG-Dataset presented in Section~\ref{section:Labeling}.}
\label{tab:table4}
\renewcommand{\arraystretch}{0.9}
\resizebox{\textwidth}{!}{%
\setlength{\tabcolsep}{8pt}
\begin{tabular}{l *{3}{c} c *{7}{c} c}
\toprule
\multirow{2}{*}{\textbf{Dataset}} & \multicolumn{3}{c}{\textbf{Labeled Gripper Type}} & \multirow{2}{*}{\textbf{Data Aug.}} & \multicolumn{8}{c}{\textbf{Success Rate(\%) $\uparrow$}} \\
\cmidrule(lr){2-4} \cmidrule(l){6-13}
& \textbf{2-finger} & \textbf{3-finger} & \textbf{4-finger} & & \textbf{WSG50} & \textbf{Franka} & \textbf{Robotiq} & \textbf{Barrett} & \textbf{Kinova} & \textbf{Delto} & \textbf{DH3} & \textbf{Avg.} \\
\midrule
(1)  & \checkmark & -          & -          & -          & 72.4 & 78.1 & 87.3 & 80.4 & 82.7 & 89.6 & 79.3 & 81.4 \\
(2)  & \checkmark & -          & -          & \checkmark & 82.7 & 79.3 & 89.6 & 88.5 & 86.2 & 86.2 & 85.0 & 85.3 \\
(3)  & \checkmark & \checkmark & -          & \checkmark & 86.2 & 80.4 & 91.9 & 90.8 & 88.5 & 88.5 & 86.2 & 87.5 \\
(4)  & \checkmark & \checkmark & \checkmark & \checkmark & 95.4 & 88.5 & 87.3 & 89.6 & 87.3 & 93.1 & 90.8 & \textbf{90.3} \\
\bottomrule
\end{tabular}}%
\vspace{-10pt}
\end{table*}
We conducted an ablation study to validate the importance of multi-gripper data augmentation in training gripper-aware models (Table \ref{tab:table4}).
Note that all experiments are evaluated on the same seven novel grippers, 
with only the training dataset composition varying across configurations.
The Jacquard dataset, which contains annotations only for the parallel-jaw gripper, serves as the baseline (1).
Configurations (2)$\sim$(4) progressively expand the XG-Dataset with multi-gripper annotations across 2-finger (2f-v1), 3-finger (3f-v1, 3f-v2), and 4-finger (4f-v1, 4f-v2) gripper types, with consistent performance gains observed across all gripper types.

The results show a consistent performance improvement as the diversity of training gripper types increases.
Using only 2-finger gripper annotations from the XG-Dataset (2) improves the average success rate from $81.4\%$ to $85.3\%$ ($+3.9\%$).
Further incorporating 3-finger gripper annotations (3) raises performance to $87.5\%$ ($+6.1\%$), 
and including all three gripper types (4) achieves the highest performance of $90.3\%$ ($+8.9\%$).
These results demonstrate that exposing the model to a broader range of gripper configurations during training significantly enhances generalization to novel grippers.

\subsubsection{Gripper aware Input Configuration}
\begin{table}[t]
\centering
\caption{Ablation study on input gripper features.}
\label{tab:table5}
\renewcommand{\arraystretch}{1.0}
\resizebox{0.75\columnwidth}{!}{%
\begin{tabular}{l cc | c}
\toprule
& \multicolumn{2}{c|}{\textbf{Input Gripper Features}} & \multirow{2}{*}{\textbf{Average Success Rate(\%)}} \\
\cmidrule(lr){2-3}
 & \textbf{Mask} & \textbf{Path} & \\
\midrule
(1) & -          & \checkmark & 73.0 \\
(2) & \checkmark &  -         & 81.4 \\
(3) & \checkmark & \checkmark & \textbf{90.3} \\
\bottomrule
\end{tabular}}
\vspace{-10pt}
\end{table}
We conducted an ablation study on the two gripper-aware input features: Gripper Mask and Gripper Path (Table \ref{tab:table5}).
Using only the Path feature (1) results in the lowest performance of 73.0\%, indicating that closing trajectory information alone is insufficient to represent gripper geometry for grasp prediction.
Using only the Mask feature (2) substantially improves performance to 81.4\%, confirming that static gripper shape plays a more dominant role in determining grasp success.
Combining both Mask and Path features (3) achieves the best performance of 90.3\%, confirming that the two features are complementary.
The Mask captures static gripper geometry while the Path encodes dynamic closing behavior, and together they provide a more complete representation of gripper-specific characteristics.

\subsubsection{Analysis of AWP Loss Designs}
\begin{table}[t]
\centering
\caption{Ablation study on AWP training strategy.}
\label{tab:table6}
\renewcommand{\arraystretch}{1.0}
\resizebox{0.75\columnwidth}{!}{%
\begin{tabular}{l | c}
\toprule
\textbf{AWP Training Strategy} & \textbf{Avg SR(\%)} \\
\midrule
MSE Loss                                        & 81.9 \\
Pairwise Contrastive Loss                       & 80.0 \\
Triplet Loss w/ Quality-aware Anchor (Ours)     & \textbf{90.3} \\
\bottomrule
\end{tabular}}
\vspace{-10pt}
\end{table}
We compare different loss function designs for AWP training to validate our selection (Table \ref{tab:table6}). 
MSE Loss yields limited generalization ($81.9\%$) as direct regression fails to incorporate gripper-aware structural constraints, preventing the model from learning the complex physical interactions across diverse geometries.  
While Pairwise Contrastive Loss separates success and failure clusters, it fails to distinguish optimal grasps from marginal ones without a quality-aware reference, resulting in a scattered embedding space ($80.0\%$). 
In contrast, our Triplet Loss with Quality-aware Anchor achieves the highest success rate ($90.3\%$). By anchoring the manifold around optimal samples, it constructs a dense, discriminative embedding space that enables robust zero-shot generalization to novel grippers.
We compare three loss function designs for AWP training to validate our design choice (Table \ref{tab:table6}).

\section{Conclusion}
We propose XGrasp, a real-time 2D planar grasp detection framework that handles diverse gripper configurations without additional training.
To address data scarcity, we constructed the XG-Dataset by augmenting existing single-gripper datasets with multi-gripper annotations based on the physical characteristics and closing trajectories of diverse grippers.
XGrasp adopts a hierarchical two-stage architecture consisting of a Grasp Point Predictor (GPP) and an Angle-Width Predictor (AWP), 
where contrastive learning with a quality-aware anchor builds a gripper-agnostic embedding space, enabling generalization to novel grippers without additional training.
Experimental results demonstrate that XGrasp outperforms existing gripper-aware methods in both grasp success rate and inference speed, achieving 90.3\% on the Jacquard dataset and 88.0\% in real-world robot experiments.
This work focuses on 2D planar grasping, which remains widely adopted in industrial settings such as bin picking and assembly due to its fast inference and practical robustness\cite{kleeberger2020survey, morrison2020learning}. 
Extending to 6-DoF gripper-aware grasping remains an open challenge, as it requires multi-gripper datasets with 6-DoF annotations. 
Future work will focus on constructing a 6-DoF XG-Dataset and designing more expressive gripper representations to extend XGrasp to gripper-aware grasp detection in 3D space.


\bibliographystyle{IEEEtran}
\bibliography{IEEEabrv,ref}

\end{document}